\documentclass{article}

\usepackage{microtype}
\usepackage{graphicx}
\usepackage{subfigure}
\usepackage{booktabs} 
\usepackage{hyperref}

\usepackage[accepted]{tom2023}
\usepackage{amsmath}
\usepackage{amssymb}
\usepackage{mathtools}
\usepackage{amsthm}

\usepackage[capitalize,noabbrev]{cleveref}

\theoremstyle{plain}

\theoremstyle{definition}

\theoremstyle{remark}

\usepackage[textsize=tiny]{todonotes}

\icmltitlerunning{The Neuro-Symbolic Inverse Planning Engine (NIPE)}

\begin{document}

\twocolumn[
\icmltitle{The Neuro-Symbolic Inverse Planning Engine (NIPE): \\ Modeling Probabilistic Social Inferences from Linguistic Inputs}





\begin{icmlauthorlist}
\icmlauthor{Lance Ying}{mit,harvard}
\icmlauthor{Katherine M. Collins}{cam}
\icmlauthor{Megan Wei}{mit}
\icmlauthor{Cedegao E. Zhang}{mit}
\icmlauthor{Tan Zhi-Xuan}{mit} \\
\icmlauthor{Adrian Weller}{cam}
\icmlauthor{Joshua B. Tenenbaum}{mit}
\icmlauthor{Lionel Wong}{mit}
\end{icmlauthorlist}

\icmlaffiliation{mit}{Massachusetts Institute of Technology, Cambridge, USA}
\icmlaffiliation{harvard}{Harvard University, Cambridge, USA}
\icmlaffiliation{cam}{University of Cambridge, Cambridge, UK}

\icmlcorrespondingauthor{Lance Ying}{lanceying@seas.harvard.edu}

\icmlkeywords{Theory of Mind, ToM}

\vskip 0.3in
]



\printAffiliationsAndNotice{}  

\begin{abstract}
Human beings are social creatures. We routinely reason about other agents, and a crucial component of this social reasoning is inferring people's goals as we learn about their actions. In many settings, we can perform intuitive but reliable goal inference from language descriptions of agents, actions, and the background environments. In this paper, we study this process of language driving and influencing social reasoning in a probabilistic goal inference domain. We propose a neuro-symbolic model that carries out goal inference from linguistic inputs of agent scenarios. The ``neuro'' part is a large language model (LLM) that translates language descriptions to code representations, and the ``symbolic'' part is a Bayesian inverse planning engine. To test our model, we design and run a human experiment on a linguistic goal inference task. Our model closely matches human response patterns and better predicts human judgements than using an LLM alone. 
\end{abstract}

\begin{figure*}[ht]
    \centering
    \includegraphics[width = 16cm]{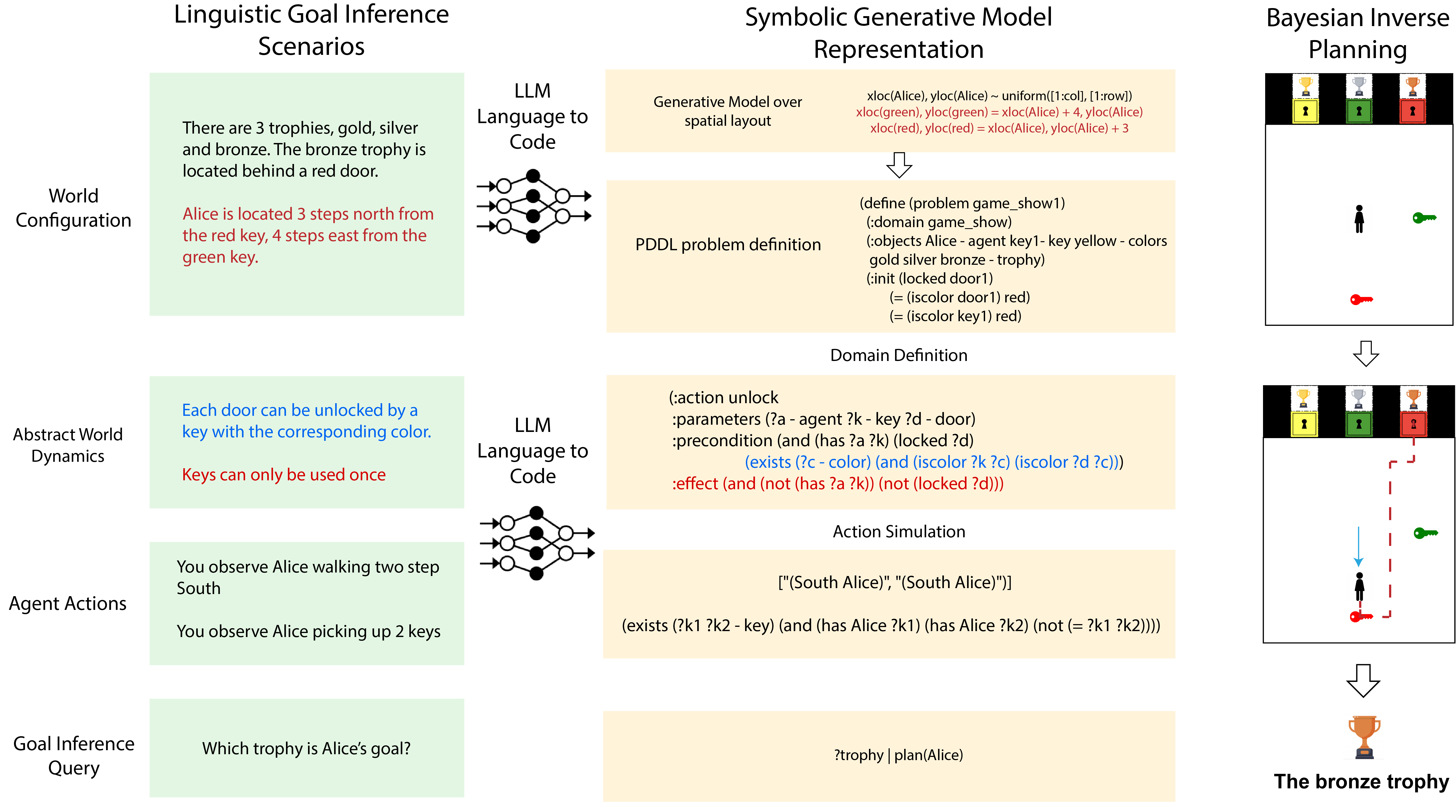}
    \caption{Overview of our model, the \textit{Neuro-Symbolic Inverse Planning Engine} (NIPE), which reasons about linguistic goal inference scenarios. (\textit{Left, green}) We design and measure human judgments on a \textit{linguistic goal inference} domain (inspired by \cite{zhi2020online}, which asks about the final goal of a participant on a complex ``gameshow" obstacle course involving trophies behind locked doors. Inferring goals requires reasoning jointly about language that describes the overall world configuration, specific world dynamics, and the agent's actions. Our model works by first using a large language model to \textit{translate} language into symbolic program semantics that collectively condition a probabilistic generative model over possible environments and agents (\textit{Middle, yellow}), and answers queries via \textit{Bayesian inverse planning} (\textit{Right}) in the conditioned probabilistic model, inferring likely goals using a nested planner to model the agent.}
    \label{fig:model}
\end{figure*}

\section{Introduction}
Humans are deeply social creatures, finely attuned to the inner lives of other people \cite{wellman1992child, saxe2006uniquely, tomasello2010origins}. We not only think, but think routinely about each other. We imagine what someone is trying to accomplish and why are trying to accomplish it; we try to parse what someone really believes and whether they might be mistaken; and we anticipate people's ultimate goals and upcoming actions, stepping in to assist or sometimes even to try to hinder them. This capacity for \textit{social reasoning}, broadly, is \textit{abstract} (we can observe someone's actions in the world, but cannot directly see what they think or want) and relies on a fine grained knowledge of distinct but highly \textit{interrelated} mental states within others (we know that someone's beliefs, desires, and goals relative to the world itself all collectively influence what they might choose to do.)

This paper considers how \textbf{language can inform social reasoning}, with a particular focus on \textbf{understanding language about people's actions, plans, and goals}. Language is an especially powerful means of conveying the abstract, interrelated nature of concepts about agents in the world---we can talk about someone's actions (\textit{I saw Annie heading into the living room}), directly convey their mental states (\textit{Annie desperately wants a toy in there}), describe relevant but abstract aspects of the world in which they act (\textit{that cupboard Annie is trying to open is locked inside}), or even pose the questions we want others to reason about (\textit{which toy do you think she wants the most?}). Language about any one of these aspects informs how we might reason about someone else in holistic ways, changing our understanding and downstream predictions about what they might know, want, or do.

Empirical evidence suggests that we think intuitively about each other as largely \textit{rational, goal-directed intelligent agents} \cite{dennett1981intentional, csibra1999goal, baker2009action, baillargeon2016psychological}. Within computational cognitive science and AI, a considerable body of prior work formalizes these intuitions by modeling agents as \textit{rational actors and planners} who choose actions to achieve specified goals or optimize rewards, and in turn models many inferences about mental states as \textit{inverse planning} from observations of actions back to their latent goals, utility functions, or beliefs about world states \cite{jara2016naive, baker2017rational,zhi2020online, ying2023infer}. Despite considerable successes in predicting distinct human inferences and modeling cooperative behavior in human-computer interactions \cite{carroll2019utility, pmlr-v119-mozannar20b, alanqary2021modeling, bansal2021most, wilder2021learning, shneiderman2022human}, these approaches---largely \textit{non-linguistic} models that draw inferences from images or demonstrations of agent actions---face scalability and specification challenges inherent to the abstract, complex, and interrelated nature of social reasoning tasks. Modeling agents as goal-directed planners requires implicitly or explicitly representing the \textit{world model} they are planning over, which is often precisely defined a priori (as in \citet{baker2017rational} and \citet{zhi2020online}) or learned from expensive exploration (as in \citet{lian2021online}). Inferring mental states, or even posing the task itself, is challenging when framed as an inductive problem from images and demonstrations alone (as in \citet{shu2021agent}).

In part due to these challenges, the recent advances made by \textit{large language models} (LLMs) have driven an explosion of interest in how these models can be used to plan \cite{valmeekam2022large} or draw social inferences \cite{bubeck2023sparks, shapira2023clever} from language---they suggest a new frontier of cognitive models and AI systems that can leverage language to specify abstract world knowledge, domain-specific world dynamics, and direct information about agents and their mental states. However, early efforts to apply LLMs directly to social reasoning problems often take a highly simplified approach to evaluating their capacity for social reasoning or their \textit{``theory of mind''}, drawing broad conclusions based on a narrow set of templated cognitive tasks (like the famous ``Sally Anne tasks''). Other evaluations caution against drawing general conclusions from such restricted task settings, pointing out that modifications and extensions beyond these simple problem settings can produce counterintuitive results that contradict the holism and generality of these inferences \cite{shapira2023clever, ullman2023large}; and several recent evaluations suggest that robust planning in general remains an outstanding challenge for large language models \cite{valmeekam2022large, bubeck2023sparks}. Several recent approaches seek to address these challenges by augmenting LLMs with interfaces to external modules for representing and simulating complex aspects of world state, particularly using classical AI representations and engines for forward planning \cite{collins2022structured, liu2023llm+, lyu2023faithful}. Emulating human-like social reasoning and knowledge, however, requires models that can robustly draw \textit{unified inferences about actions, mental states, and the world} \cite{wong2023word}, relating plans and beliefs in holistic, flexible, and ideally interpretable and safely deployable way to language; and linguistic evaluations that move in turn towards the richness and breadth of human-like social reasoning, flexible and extensible enough to incorporate these interrelated aspects of human mental states and world knowledge.

We focus specifically in this paper on \textbf{understanding language that describes inverse planning and goal inference problems}---we consider how language can convey information about the world itself, and the actions within it, in order to drive inferences about the unseen goals motivating these actions. We make the following contributions: 
\begin{enumerate}
    \item We introduce a new model, the \textit{Neuro-symbolic Inverse Planning Engine} (NIPE) that leverages LLMs to \textit{translate language into program expressions} that compositionally construct and condition a \textit{probabilistic generative model over agents and environments}, then uses \textit{Bayesian inverse planning} to perform goal inference over this symbolic representation, expanding upon the paradigm of \emph{rational meaning construction} for language-informed cognition \cite{wong2023word}.
    \item To evaluate this model, we compare our model against \textit{human social reasoning on a linguistic inverse planning domain} inspired by prior cognitive work in social cognition \cite{zhi2020online}. We design this domain to require jointly reasoning about agent actions and varying information about the world itself.
    \item We find that our model \textit{well-predicts human judgments} across a variety of distinct social reasoning scenarios. We also find that our model \textit{outperforms LLM baselines used to directly perform the same goal inference tasks}. We find that the largest LLM alone (GPT-4 \cite{openai2023gpt4}) correlates moderately well with certain human judgments on this domain, particularly on simpler scenarios. We discuss trends between LLMs of different sizes and on different task variants in our stimuli.

    
\end{enumerate}
\begin{table*}[ht!]
    \centering
    \footnotesize
    \begin{tabular}{@{}p{0.25\linewidth}p{0.7\linewidth}l@{}}
    
        \toprule
        \multicolumn{2}{l}{\textbf{Linguistic planning and inverse planning concepts across tasks}} \\ \midrule
        
        \textbf{\textit{World configuration}} & There are three trophies placed through this obstacle course: gold, silver, and bronze; The gold trophy is currently 5 steps East of Alice; The silver trophy is behind 1 door. \\ \\

        \textbf{\textit{World dynamics}} & On this course, each lock can only be opened by a key that is the same color as that lock; In this game, players lose points if they pick up more keys than needed. \\ \\

        \textbf{\textit{Agent actions}} & You see Alice picking up a green key; You see Alice walking a step towards the South; In this game, players lose points if they pick up more keys than needed. \\ \\

        \textbf{\textit{Goal inference}} & Considering each of the possible trophies on this course, how likely is it that Alice is trying to reach this trophy? \\

         \midrule
        \multicolumn{2}{l}{\textbf{Task variations}} \\ \midrule
        
        \textbf{\textit{Generic Keys}} & There are three trophies placed throughout this obstacle course: gold, silver, or bronze. To get to the gold trophy, you need to unlock 2 different doors. To get to the silver trophy, you need to unlock 1 door. The bronze trophy isn't behind a door. Keys to unlock the doors are placed around the obstacle course. \textbf{All of the keys on this course work on all of the locks.} You see Alice pick up 1 key.\\ \\

        \textbf{\textit{Same-Colored Keys}} & There are three trophies placed throughout this obstacle course: gold, silver, or bronze. The gold and silver trophies are both located behind a yellow door. To go to the bronze trophy, you need to go through a red door. On this course, \textbf{each lock can only be opened by a key that is the same color as that lock}. You see Alice pick up a red key.\\ \\
       \textbf{\textit{Different-Colored Keys}} & There are three trophies placed throughout this obstacle course: gold, silver, or bronze. The gold and silver trophies are both located behind a green door. To go to the bronze trophy, you need to go through a red door. On this course,\textbf{ each door can be unlocked by any key that is a different color than the door}. You see Alice pick up a red key.\\ \\ 

        \textbf{\textit{Spatial}} & There are three trophies placed throughout this obstacle course: gold, silver, or bronze. Alice is a participant on this course. The gold trophy is currently \textbf{5 steps East of Alice}. The silver trophy is \textbf{3 steps South from her}. The bronze trophy is \textbf{5 steps further West of the silver trophy}. You see Alice walking a step towards the South.\\
        \bottomrule
    \end{tabular}
    \caption{Example language from the linguistic ``gameshow" goal inference domain. (\textit{Top}) Each scenario contains language that spans a variety of interrelated concepts relevant to the final goal-inference query, ranging from information about the particular environment configuration to varying information about an agent's actions. (\textit{Bottom}) We design four distinct task variants that vary in complexity and type of the environment description and the spatial details of the agent's actions.}
    \label{tab:stimuli}
\end{table*}

\section{Linguistic inverse planning tasks and human experiment}
We begin by describing the domain of tasks and human experiment we use throughout this paper, to provide intuition for the modeling experiments that follow. We construct a domain of \textbf{linguistic inverse planning stimuli} designed to evaluate how humans and models \textbf{infer the goals of agents acting within varied environments}. We design stimuli to convey varying information about the \textit{world configuration and dynamics} in which agents behave, as well as the \textit{observed actions of agents} with unknown goals. 

Our tasks are inspired by the spatial navigation stimuli used in prior social reasoning experiments \cite{zhi2020online,alanqary2021modeling}. Our experiments adapt this basic domain into a set of linguistic inverse planning stimuli, and introduce a range of variations designed to test distinct world dynamics, spatial information, and vague or incomplete information about agent actions conveyed linguistically.

\subsection{Planning and inverse planning concepts in the Gameshow Domain}
Each stimuli in our experiment (\cref{tab:stimuli}) begins with a description of a ``\textit{gameshow}" scenario in which a player is trying to reach one of several goal \textit{trophies} placed around an obstacle course. This obstacle course also includes \textit{locked doors} that can only be unlocked with keys placed around the course, and which additionally require multi-step reasoning about an agent's longer term plans. Based on each scenario description, which includes varying information about the particular obstacle course and the actions of an agent within it, we pose a linguistic goal inference query: \textit{which of the possible trophies do you think the player trying to reach?} 

We design each stimulus to span several core conceptual categories relevant to reasoning about agents and their actions in the world (\cref{tab:stimuli}, \textit{top}), including:
\begin{itemize}
\itemsep0em
    \item \textbf{World configuration}: information about goal trophies, keys, and their attributes, as well as their relative locations, at differing levels of specificity.
    
    \item \textbf{Abstract world dynamics}: causal dynamics within the environment, such as rules determining which keys can open which doors.

    \item \textbf{Agent actions}: descriptions of an agent's behavior before they reach the goal, like \textit{picking up keys} or \textit{walking} in different directions towards keys or trophies.
\end{itemize}

\subsection{Task variants}
We design four distinct variants of gameshow stimuli for a total of \textit{n=18} stimuli, which vary the complexity of the specified world configuration, the world dynamics, and the type of observed agent actions (excerpts from stimuli in each variant are shown in \cref{tab:stimuli}, \textit{bottom}):

\begin{itemize}
\itemsep0em
    \item \textbf{Generic keys:} Goal trophies are placed behind locked doors, and must be unlocked by keys placed around the obstacle course. Any key on the course can be used to open any other door.
    \item \textbf{Same-colored keys:} This variant adds additional complexity in the world dynamics. Keys and doors have colors. Doors can only be unlocked by keys that are the same color as the door.
    \item \textbf{Different-colored keys:} In this variant, doors can only be unlocked by keys that are a \textit{different} color from the door.
    \item \textbf{Spatial:} This variant contains specific spatial descriptions of the obstacle course and agent actions. Due to the added spatial complexity, in these environments, there are no keys or doors; agents can walk freely towards various goal trophies.
\end{itemize}

\subsection{Human experiment}
We collect and evaluate human judgments on these goal inference tasks. We recruit a total of n=46 participants from Prolific \footnote{We exclude 6 participants as less than 30\% of their MAP results agreed with consensus; i.e. they disagree with the majority on which goal is the most likely for at least 5 out of 7 stimuli.}; each viewed a random batch of 7 stimuli from a specific task variant, and for each stimulus, produced goal inference judgements about each possible goal trophy on a 1-7 Likert scale (\textit{1: definitely not this trophy; 7: definitely this trophy}). Scenarios were presented for a fixed amount of time before an answer was permitted (90 seconds in the \textit{Spatial} condition; 45 seconds in the others). Each scenario was seen by between 14-16 participants. Participants were native English speakers from the USA/UK.

\section{Our Model: The Neuro-Symbolic Inverse Planning Engine (NIPE)}
We now introduce our model, the \textit{Neuro-Symbolic Inverse Planning Engine} (NIPE). While we focus on goal inference tasks, we describe a general framework for conditioning on the diverse range of linguistic information in these stimuli---varying information about the spatial environment, world dynamics, and agent actions---that supports inferences about agents based on a principled, probabilistic framework for Bayesian inverse planning. 

Our core contribution is to show how we can approach complex \textit{linguistic} scenarios about agents by leveraging structured, symbolic representations and inference methods (modeling agents as explicit \textit{planners}, and performing principled Bayesian inference over complex models) from the existing cognitive and AI social reasoning literature (eg. \cite{baker2017rational,zhi2020online}). We do so by \textbf{translating from language into program expressions} that condition a \textbf{symbolic generative model over agents and environments}, instantiating the paradigm of \emph{rational meaning construction} for language-guided cognition \cite{wong2023word}. This approach also addresses an important open challenge for prior non-linguistic (inverse) planning models: We can condition on abstract information about the world that is specified in language, such as information about \textit{where objects are located} or \textit{which keys open which doors}.

Our model (shown in schematic form in \cref{fig:model}) modularly decomposes the problem of goal inference from language into three general subproblems:
\textit{language-to-code translation}, which we instantiate generally by using a LLM as a few-shot semantic parser (similar to  \cite{zhang2023grounded,liu2023llmp,collins2022structured, wong2023word}); the construction of a \textit{probabilistic generative model over agents and environments} from language; and \textit{goal inference through Bayesian Inverse Planning} in this probabilistic model (using an nested planning and inference algorithm like that in \citep{zhi2020online}). We now describe each problem in detail.

\subsection{Language-to-code translation using LLMs} 
Following \citet{wong2023word}, the crux of our approach is to show how the different kinds of language in our scenarios can each be translated into program expressions that collectively construct and condition a holistic probabilistic model. We begin, however, by briefly describing the language-to-code translation itself, which we implement by using LLMs to \textit{translate sentences into code}, which compositionally construct a symbolic model over an extended, multi-sentence linguistic scenario (see \cref{fig:model}, \textit{LLM Language to Code}).

Our translation approach leverages the few-shot prompting capabilities of LLMs trained jointly on language and code. We choose \textit{GPT-3.5} \citep{ouyang2022training} specifically to show that a \textit{smaller} LLM can be used to implement this more restricted task, rather than using the LLM itself to internally represent and reason over these scenarios. Future work should explore whether even smaller and more restricted language-to-code models can be used. Our approach builds on the methods used in \citet{zhang2023grounded}, which uses an LLM to translate language into probabilistic programs, along with other work showing that LLMs can translate language into code when given paired few-shot examples \citep{liu2023llm+,wong2023word}. For each stimulus, we construct a prompt containing 2-3 randomly sampled examples of paired language and code from held out examples on our domain. Models were sampled with a temperature of $1.2$. To ensure robustness, we rejection sample translations to ensure that they pass a number of syntactic and semantic tests (e.g., to ensure that translated code is valid and executable.)

\subsection{Constructing a probabilistic generative model over agents and environments from language}
Now we describe how the different kinds of information relevant to this domain can be each translated into different kinds of code within a unifying generative model. Following \citet{zhi2020online}, we model information about the world dynamics, such as action preconditions and effects, using the Planning Domain Definition Language (PDDL), a standardized representation language widely used in robotics and automated planning \cite{aeronautiques1998pddl}, and which can b extended with modular semantics to efficiently represent spatial domains such as ours \cite{gregory2012planning,zhi2022pddl}. 

While prior work has considered how different subproblems related to planning can be extracted via language-to-code \citep{collins2022structured,liu2023llmp, silver2023generalized}, our aim here is to translate language not only to distributions over possible worlds 
--- i.e. possible spatial layouts and domain rules --- but also to inferences over an agent's actions. We describe each aspect of translation below:

\paragraph{Conditions on spatial environments}
First, we prompt LLM to translate sentences like ``The bronze chest is located behind a red door.'' into a \texttt{CONDITION} statement that describes this arrangement. We then sample concrete grid-world maps from a generative model over maps that satisfy this.

To ensure that maps are valid for our environment, we define and condition a relatively restricted generative model over possible maps and trophy placements that ensures that these basic validity conditions are met.
Each sampled map from this generative model is a concrete PDDL scene representation that describes the locations of the initial agent, objects, and their attributes.

\paragraph{Conditions on domain rules}
Now, consider sentences like ``Each door can be unlocked by a key with the corresponding color.'' These can be straightforwardly translated into operator definitions expressed in PDDL (see Figure \ref{fig:model}), which define an action and its necessary parameters, preconditions for trying this action, and effects on the environment.

Collectively, the scene representation and the domain rules define a complete, concrete planning domain defined in PDDL. As in \citet{zhi2020online}, this domain model is nested within a larger Bayesian model over agents acting in this environment. The benefit of using PDDL is that we can easily model the agents themselves using classical planning algorithms over the PDDL domain representation and we can easily transfer this model to a different problem setup by modifying the operator definitions or simply choosing a different pre-defined PDDL domain.

\paragraph{Conditioning a generative model over agent subgoals and actions from language}

Finally, we consider how to translate sentences that describe agents' actions into conditions on a model over agents. We distinguish between language like ``You observe Alice picking up two keys'' which is treated as a condition on subgoals, and ``You observe Alice walking 3 steps East.'', which is treated as a low-level action sequence.

\subsection{Goal Inference through Bayesian Inverse Planning}

Once we have an initial PDDL environment representation $s_0$ and a sequence of agent actions $a_{1:t}$, we can perform Bayesian inverse planning to infer a posterior distribution over goals $g$ given actions $a_{1:t}$ and the initial state $s$, which is proportional to the joint distribution over goals and actions:
\begin{equation*}
    P(g | s_0, a_{1:t}) \propto P(g, a_{1:t} | s_0) =
        P(g | s_0) P(a_{1:t} | g)    
\end{equation*}
Here $P(g | s_0)$ defines the prior distribution over possible goals in the initial state $s_0$, and $P(a_{1:t} | g)$ is a model of how the agent selects actions $a_{1:t}$ given a particular goal $g$. For the goal prior, we model  agents as being more likely to pick up goals that are closer to them, as determined by the cost $C(g, s_0)$ of the shortest plan to each goal:
\begin{equation*}
    P(g | s_0) \propto \textstyle{\frac{1}{C(g)}}
\end{equation*}
To model actions, we assume $P(a_{1:t} | g)$ decomposes into the probability of selecting an action $a_t$ at each state $s_t$, given by a Boltzmmann-rational policy $\pi$:
\begin{align*}
P(a_{1:t} | g) &= \textstyle{\prod}_{\tau=1}^t \pi(a_\tau | s_t, g) \\
\pi(a_t | s_t, g) &= \frac{\exp{\beta Q_g(s_t, a_t)}}{\sum_{a_t'} \exp{\beta Q_g(s_t, a_t')}}
\end{align*}
where $Q_g(s_t, a_t) $ denotes the (negative) cost of the reaching the goal $g$ by taking action $a_t$ from $s_t$, and $\beta$ is a rationality parameter, also called the inverse temperature. This policy models agents as approximately rational in how they act, since actions which reach the goal more quickly (higher $Q_g$) are more probable, with higher values of $\beta$ leading to more optimal actions. Although this model does not account for forms of approximate rationality \cite{evans2016learning,shah2019feasibility,alanqary2021modeling}, it is a widely used choice in (Bayesian) inverse planning and inverse reinforcement learning that can account for non-systematic deviations from optimality \cite{ramachandran2007bayesian,ziebart2008maximum,baker2009action}. To compute $Q_g$ for each state $s_t$ and action $a_t$ encountered during inference, we use A* search as an optimal-cost planner in the PDDL domain to determine the cost of the optimal plan.

With these components of the model specified, goal inference can be performed \emph{exactly} by computing the joint probability $P(g, a_{1:t} | s_0)$ of each possible goal $g$ with the observed sequence of actions $a_{1:t}$, which can then be normalized to obtain the posterior distribution $P(g | s_0, a_{1:t})$.

\section{Model Experiments}
We compare our model with human judgements on the gameshow domain of social reasoning tasks. We also evaluate two LLM baselines to assess the value of an explicit Bayesian inverse planning model.

For our model, we simulate answers to each goal inference stimulus (\textit{which trophy will Alice pick up?}), outputting a \textit{posterior probability} over all of the possible trophies (probabilities sum to 1 over all trophies). For each stimulus, we translate the linguistic scenario description of the world configuration, world dynamics, and agent behavior into symbolic conditions on the generative model (by sampling from the LLM and rejecting invalid code), and perform goal inference with the generated code to calculate the posterior over the trophies. Since goal inference can be performed without sampling, and since the space of maps consistent with a linguistic description is highly restricted, we found in pilot experiments that our model showed limited variability across LLM and generative model samples. Therefore, for each stimulus, we computed our results using a single valid sample of PDDL domain rules and spatial environments.

We also conduct two baseline experiments using two different LLMs (\texttt{gpt-3.5-turbo-0301}, the dialogue-tuned variant of GPT-3.5~\citep{ouyang2022training}, and \texttt{gpt-4-0314}, the March snapshot of GPT-4~\citep{openai2023gpt4}) to directly provide goal inference judgments for each stimulus. We assess LLMs using exactly the same conditions as the human experiment, using a zero-shot prompt where the LLM is given the full linguistic experimental set up (including instructions explaining the task) shown to human participants, concatenated with the scene scenario and query for each stimulus, and a short prompt specifying the answer format. To account for answer variability, we sample \textit{n = 30} answers for each stimulus, rejecting any answers that are not in the correct format. 
Exact prompt formats can be found in the Appendix.
\section{Results and discussion}

\begin{figure*}[t!]
    \centering
    \includegraphics[width=16cm]{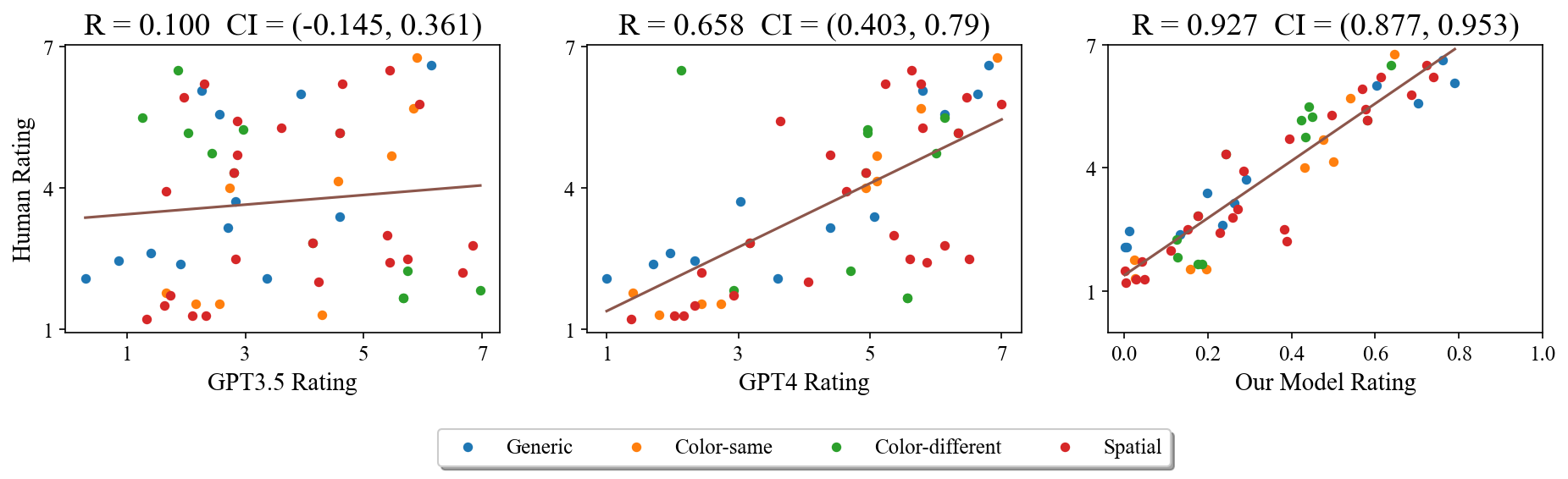}
    \caption{Correlation between NIPE and baseline LLMs vs. human judgments, aggregated across the goal-inference stimuli. Our model (\textit{Right}) produces posterior probabilities over goals that correlate strongly overall with human judgments; GPT-4 correlates moderately overall (\textit{Middle}) and GPT-3.5 (\textit{Left}) correlates poorly across the tasks. Color-coded points show stimuli from specific task variants.}
    \label{fig:corr_all}
\end{figure*}

\begin{figure*}[ht!]
    \centering
    \includegraphics[width=16cm]{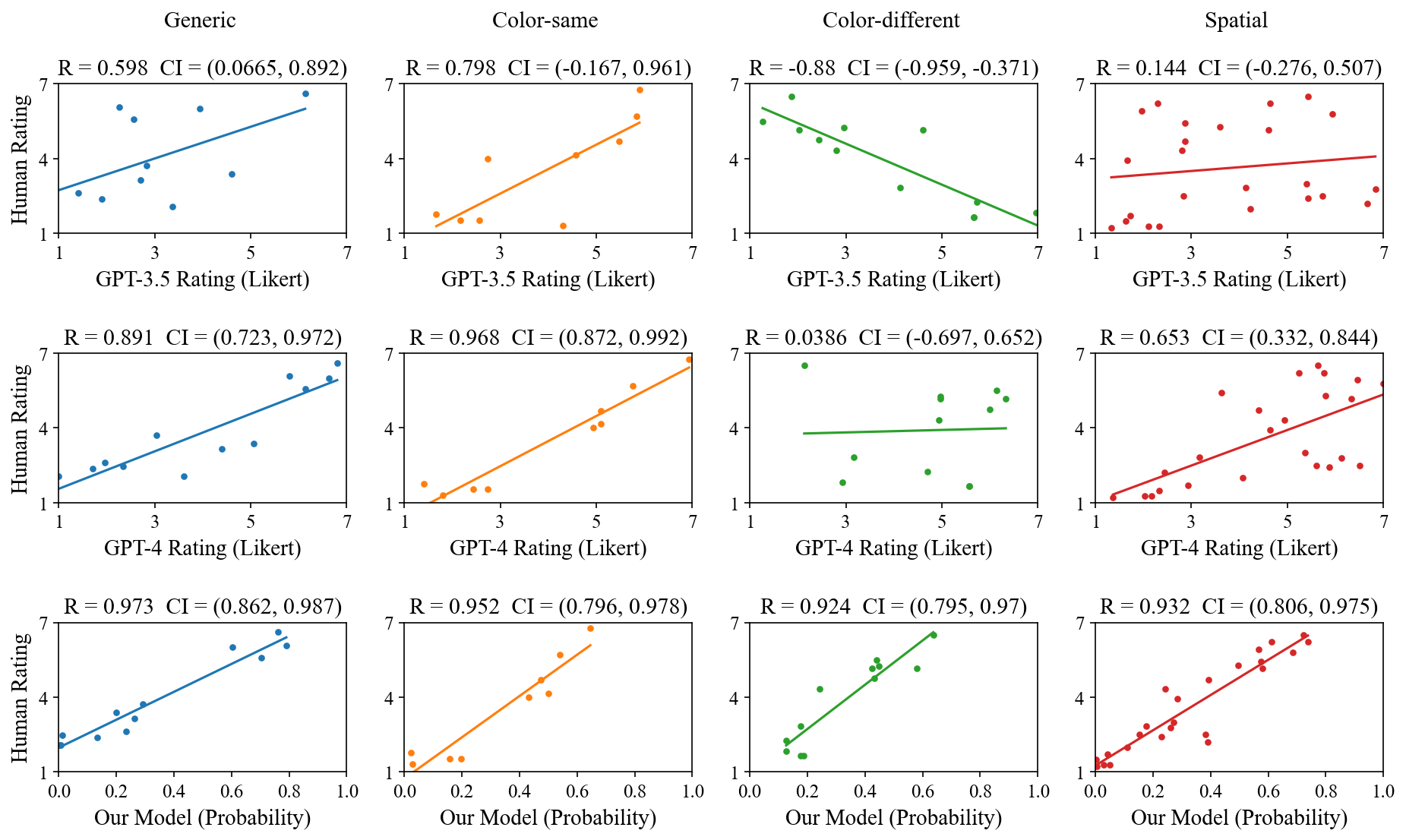}
    \caption{Correlations within specific task variants highlight more granular trends in model behavior. Our model (\textit{Bottom}) correlates strongly with human judgements across all four task variants. In comparison, the LLM baselines generally show poorer performance on the more complex variants (the \textit{Color-Different} variant, which introduces a new, counterintuitive rule about the world dynamics; and the \textit{Spatial} variant, which requires reasoning specifically about agents as efficient path planners.}
    \label{fig:corr-categories}
\end{figure*}

We evaluate our model and baselines in comparison to human judgments, both across all of the stimuli (\cref{fig:corr_all}) and within each task variant category (\cref{fig:corr-categories}). We find that:
\paragraph{Our model (NIPE) correlates strongly overall with human judgments across the goal inference experiment.} We calculate correlations between the \textit{posterior probability judgments} produced by our model and the mean per stimulus judgments (Likert scale ratings) across human subjects, and find that our model is strongly correlated with human judgments in the experiment overall (\cref{fig:corr_all}, \textit{Our Model}, R=0.927, 95\% CI=(0.877, 0.953)). All 95\% CI are bootstrap sampled confidence intervals each with 1000 samples.

\paragraph{The largest LLM baseline (GPT-4) correlates moderately with human judgments overall; GPT3.5 correlates poorly.} We additionally calculate correlations between LLMs prompted directly to produce goal inference judgments (mean over sampled LLM judgements for each stimulus, and the mean over human judgments for each stimulus across participants). In comparison to our model, we find that GPT-4 correlates well, but more weakly, with human judgments (\cref{fig:corr_all}, \textit{GPT-4 Rating}, R=0.658, CI=(0.403, 0.79)) overall, and GPT-3.5 correlates much more poorly with human judgments overall  (\cref{fig:corr_all}, \textit{GPT-3.5 Rating}, R=0.100, CI=(-0.145, 0.361)). These aggregate trends are better understood by examining correlations between human judgments within each task variant, as we discuss next.

\paragraph{NIPE correlates strongly with human judgements \textit{within} each task variant; GPT-3.5 and GPT-4 are less robust to complex world dynamics and spatial information.} Inspecting trends \textit{within} each individual task variant (\cref{fig:corr-categories}) provides insight into when models deviate from humans.

NIPE consistently correlates with human judgments across \textit{all} task variants, from the simpler \textit{Generic} stimuli to the more complex \textit{Colored} and \textit{Spatial} stimuli (\cref{fig:corr-categories}, bottom). In contrast, on the more complex tasks---particularly the \textit{Colored-Different} variant, which specifies a new, counterintuitive rule particular to this domain (\textit{doors can only be opened by keys that are a \textbf{different} color from the door}), and the \textit{Spatial} variant, which requires considering how an efficient, path-planning agent would navigate towards their goal---the LLM baselines show revealingly different performance in comparison to human reasoning. 

As seen in \cref{fig:corr-categories}, GPT-3.5 actually correlates \textit{negatively} with human judgments on \textit{Colored-Different}---it both ignores the stated rule about the different colored keys and appears to make the opposite, incorrect assumption, perhaps based on simpler color-based pattern matching; and shows nearly zero correlation with human judgments on the explicit \textit{Spatial} path-planning variant. In comparison, GPT-4 correlates well with human judgments on both of the simpler task variants, and shows a moderate (but positive) correlation on the \textit{Spatial} variant (though we discuss stimuli-specific failures next, suggesting that GPT-4 may still be using simpler distance heuristics rather than robustly modeling an efficient path-planning agent.) On the \textit{Colored-Different} domain, GPT-4 has nearly zero-correlation, suggesting that it is highly uneven in taking the specific rule about key colors into account---as shown in \cref{fig:corr-categories}, it predicts human judgments well in some cases, but produces nearly the opposite overall judgment in others, suggesting it has ignored the rule.

\paragraph{Humans make fine-grained assumptions about agents as rational planners; LLMs do not always do so.} Our model explicitly reasons about agents as rational, goal-directed actors and planners. Do LLMs do so as well? Qualitative analysis of individual stimuli highlights subtle but important discrepancies between human assumptions and LLM-based goal inferences on the most complex \textit{Spatial} stimuli. We highlight a representative example here: 
\vspace{-4ex}
\begin{quote}
   \footnotesize
   \textit{There are three trophies placed throughout this obstacle course: gold, silver, or bronze. Alice is a participant on this course. The gold trophy is currently 2 steps East of Alice. The silver trophy is 3 steps South from her. The bronze trophy is 5 steps South from her. You see Alice walking 4 steps South.} 
\end{quote}
\vspace{-2ex}
Both humans and our model strongly predict that Alice is headed towards the \textit{bronze} trophy on this domain. In contrast, both the LLM baselines predict most strongly that Alice is heading towards the \textit{silver trophy} (perhaps because it is closer to Alice), though in fact, humans tend to agree that because she has walked \textit{past} this trophy, as an efficient actor, this is likely not her goal. We provide additional examples of similar efficient-planning situations in the Appendix.

\section{Conclusion and Future Directions}

We conclude with several important avenues for future work. First, while we evaluate all LLMs using a simple prompting regime, future work should evaluate whether these results are consistent or improve with other prompting and evaluation methods such as chain-of-thought \citep{wei2022chain, kojima2022large}, particularly on complex tasks such as those that invoke efficient spatial path planning, which prior work also finds to be an outstanding challenge for LLMs \citep{cohn2023dialectical}. It is also possible for LLMs to be biased by surface-level cues (e.g. in our stimuli, they might be biased towards gold over silver trophies, since the former are often considered more valuable). Future experiments should control for these potential biases by testing a wider diversity of prompts. 

Second, while NIPE adopts a Bayesian approach to social inference, the process of language translation itself is performed in a non-Bayesian ``bottom-up'' manner using LLMs. Future work might explore a more fully Bayesian approach that models how language itself is generated from symbolic task representations, and perform posterior inference directly from language. This might involve using LLMs as likelihood models over language, as in \citet{ying2023infer}, and also incorporate LLMs as common sense priors over representations of the world \citep{lew2020leveraging,li2023lampp}. Among other benefits, this could enable more controlled and robust translation of language to code \cite{poesia2022synchromesh, lew2023sequential}.

Finally, we designed the gameshow domain to permit future extensions that describe a much broader range of social reasoning tasks that we could express linguistically. This includes scenarios that specifically invoke agent belief \cite{baker2017rational,ullman2023large}; involve boundedly rational agents \cite{alanqary2021modeling}; and even involve multiple agents collaborating or competing \cite{kleiman2016coordinate}. In addition, future work could extend this domain to multimodal stimuli that include both linguistic information \textit{and} demonstrations or images, e.g., to convey complex agent trajectories or environment layouts \cite{ying2023infer}. The difficulty of the domain could also be increased, expanding the number of objects, actions, and planning steps in order to push the limits of language-to-code translation. By building out this architecture, we hope to fully reap the benefits of the both the neural and symbolic traditions of AI, enabling flexible social inferences from a wide range of naturally occurring linguistic stimuli, while producing outputs that are structured, coherent, and robust \cite{collins2022structured}.

\bibliography{example_paper}
\bibliographystyle{tom2023}

\newpage
\appendix
\onecolumn
\section{Appendix}

\subsection{Acknowledgments}

TZX is funded by an Open Phil AI Fellowship. KMC is funded by a Marshall Scholarship and the Cambridge Trust. AW acknowledges support from a Turing AI Fellowship under grant EP/V025279/1, The Alan Turing Institute, and the Leverhulme Trust via CFI. The work is additionally funded by DARPA, MCS, AFOSR, ONR Science of AI programs, and the MIT-IBM Watson AI lab.

\subsection{Human linguistic inverse planning experiment}
We include additional details on our human experiment. For all scenarios, participants were asked ``Now, considering each of the possible trophies on the course, please respond on the 1-7 scale: how likely is it that the participant is trying to reach this trophy?'' Likert scale options were as follows: 

\begin{itemize}
    \item ``1 (Definitely not this trophy)''
    \item ``2 (Very unlikely)''
    \item ``3 (Somewhat unlikely)''
    \item ``4 (Equally likely or unlikely)''
    \item ``5 (Somewhat likely)''
    \item ``6 (Very likely)''
    \item ``7 (Definitely this trophy)''
\end{itemize}

Note, due to a mismatch, some participants in the \texttt{Generic} condition saw one stimuli from \texttt{Different-Colored Keys} and vice versa. We reversed fixed this partway through the study.



\subsection{LLM Baselines Experiment}

We use the gpt-4-0314 model (i.e., a fixed, non-updatable checkpoint from March 14 for maximal reproducibility) in the Chat Completion API. For each stimuli, we sample until we get 20 well-formed responses, rejecting samples that are not in the correct format.

\textbf{Example LLM Prompt for World Configuration}
\begin{verbatim}
  Input:
  There are three trophies placed throughout this obstacle course: gold, silver, or bronze. 
  The gold trophy is behind a green door. 
  The silver trophy is behind a yellow and red door. 
  The bronze trophy is behind a red door. 
  
  Keys to unlock the doors are placed around the obstacle course. On this course, each door 
  can be unlocked by any key as long as it is of a different color than the door.
  
  You see Alice picking up a red key. What would be the likely trophy Alice wants to get?"


  output:
  {"agent" : ["Alice"],
  "goals" : ["gold", "silver", "bronze"], 
  "locations" : {"gold" : "Room A", "silver" : "Room B", "bronze" : "Room C"},
  "obstacles" : {"Room A" : ["G"], "Room B" : ["Y", "R"],"Room C" : ["R"] },
  "keys" : ["g","y","r"], 
  "max_obstacle" : 2,
  "keys_per_door": 1,
  "len_key" : 3,
  "goal_count" : 3,
  "observation_type": "has_objects",
  "observation": "(exists (?k - key) (and (has Alice ?k) (iscolor ?k red)))"}

\end{verbatim}

\textbf{Example LLM Prompt for PDDL Domain Definition}

    In this example we constrain the model such that the LLM can only modify the unlock action of the PDDL domain definition. Our goal is to show that with few-shot prompting in the color-same condition, the LLM can generalize to generate action definition for color-different condition.
    
\begin{verbatim}

    input:
    
    Alice is playing a treasure game to pick up exactly one trophy. 
    There are 3 trophies: gold, silver and bronze, located in room A, room C and room B. 
    There is 1 yellow and 1 green lock for Room A, a red lock for room B and a 
    yellow lock for room C. 
    
    Alice needs to unlock the locks to enter the rooms. Each lock can be unlocked
    with a key with the same color. Each key can only be used once.
    
    You see Alice go over to pick up a red key.
     
    It's unclear what Alice would do next.

    output:

    (:action unlock
     :parameters (?a - agent ?k - key ?d - door)
     :precondition (and (has ?a ?k) (locked ?d)
                        (exists (?c - color) (and (iscolor ?k ?c) (iscolor ?d ?c)))
                        (or (and (= (xloc ?a) (xloc ?d)) (= (- (yloc ?a) 1) (yloc ?d)))
                            (and (= (xloc ?a) (xloc ?d)) (= (+ (yloc ?a) 1) (yloc ?d)))
                            (and (= (- (xloc ?a) 1) (xloc ?d)) (= (yloc ?a) (yloc ?d)))
                            (and (= (+ (xloc ?a) 1) (xloc ?d)) (= (yloc ?a) (yloc ?d)))))
     :effect (and (not (has ?a ?k)) (not (locked ?d)) )
    )
\end{verbatim}

\subsection{An example of LLM baseline result}
Here we provide an example of GPT-4's response in the baseline condition on the \textit{Color-different} task variant.

\begin{verbatim}

    input:

    There are three trophies placed throughout this obstacle course: 
    gold, silver, or bronze.
    
    The gold and silver trophies are both located behind a yellow door. 
    To go to the bronze trophy, you need to go through a red door.
    
    Keys to unlock the doors are placed around the obstacle course.  
    On this course, each door can be unlocked by any key that is a different 
    color than the door. For example, yellow keys cannot unlock yellow doors, 
    but can unlock doors in green or red.
    
    In this game, players lose points if they pick up more keys than needed. 
    You may assume that players will use all the keys they pick up.
    
    Alice is a participant on this course. You see Alice pick up a yellow key.

    It's unclear what Alice would do next.

    How likely is Alice picking up the trophies? Please rate from 1 to 7 
    and explain.

    output:

    Gold Trophy: 6
    Silver Trophy: 6
    Bronze Trophy: 1

    Explanation:

    Since Alice has picked up a yellow key, she can now unlock either 
    the gold or silver trophy, since both are located behind a yellow 
    door. This makes it quite likely that she will pick up one of those
    two trophies (rating them both at 6).

    However, picking up a yellow key means she cannot unlock the red
    door, which leads to the bronze trophy. Therefore, it is very unlikely
    that Alice will pick up the bronze trophy (rating it 1).
    
\end{verbatim}

It is clear from this example that GPT-4 shows unsatisfactory understanding of the setup of the \textit{Color-different} variant. It is stated that yellow keys cannot unlock yellow doors, so it is unlikely that Alice would pick up the gold or silver trophy, but GPT-4 still judges these two possibilities to be highly likely. On the other hand, both humans and our NIPE model assign significantly more likelihood to the bronze trophy (and roughly equal low likelihoods to the gold and silver trophies) in this example.

\subsection{Sample stimuli for each experimental condition}
\textbf{Generic:}
\begin{quote}
There are three trophies placed throughout this obstacle course: gold, silver, or bronze.   

To get to the gold trophy, you need to unlock 2 different doors. To get to the silver trophy, you need to unlock 1 door. The bronze trophy isn't behind a door. 

Keys to unlock the doors are placed around the obstacle course. On this course, each door has exactly 1 lock on it. All of the keys on this course work on all of the locks. However, each key can only be used one time.

Alice is a participant on this course. You see Alice go over and pick up two keys.
\end{quote}

\textbf{Color-same:}
\begin{quote}
There are three trophies placed throughout this obstacle course: gold, silver, or bronze.   

The gold and silver trophy trophies are both located behind a yellow door. To get to the yellow door, you first need to go through a green door. 
To go to the bronze trophy, you need to go through a red door. 

Keys to unlock the doors are placed around the obstacle course. On this course, each door is unlocked by a key of the corresponding color.
 
In this game, players lose points if they pick up more keys than needed. You may assume that players will use all the keys they pick up.

Alice is a participant on this course. You see Alice pick up a yellow key.
\end{quote}
\textbf{Color-different:}
\begin{quote}
There are three trophies placed throughout this obstacle course: gold, silver, or bronze, which are placed in rooms labeled A, B, and C, respectively.  

 Room A is behind a door with a yellow and green lock on it. 
 Room B is behind a door with a red lock on it. 
 Room C is behind a door with its own yellow lock on it. 

Keys to unlock the doors are placed around the obstacle course.  On this course, each door can be unlocked by any key that is a different color than the door. For example, yellow keys cannot unlock yellow doors, but can unlock green and red doors 

In this game, players lose points if they pick up more keys than needed. You may assume that players will use all the keys they pick up.

Alice is a participant on this course. You see Alice pick up a yellow key.
 
It's unclear what Alice would do next.
\end{quote}
\textbf{Spatial:}
\begin{quote}
There are three trophies placed throughout this obstacle course: gold, silver, or bronze. Alice is a participant on this course.  

The gold trophy is currently 1 step East of Alice. 
The silver trophy is 3 steps South from her. 
The bronze trophy is 3 steps East from the silver trophy. 

You see Alice walking 3 steps East.
\end{quote}
\end{document}